# Low Cost Swarm Based Diligent Cargo Transit System

Harish K[1], Varadhan R[2], Anurag R M[3], Harmanpreet S[4]
[1,2,3] National Institute of Technology Trichy, [4] Intel India
Karunaha73925@gmail.com[1], varadhan1995nitt@gmail.com[2], anuragmurali94@gmail.com[3],
harmanpreet.s.bambah@intel.com[4]

*Abstract*— **The goal of this paper is to present the design and development of a low cost cargo transit system which can be adapted in developing countries like India where there is abundant and cheap human labour which makes the process of automation in any industry a challenge to innovators. The need of the hour is an automation system that can diligently transfer cargo from one place to another and minimize human intervention in the cargo transit industry. Therefore, a solution is being proposed which could effectively bring down human labour and the resources needed to implement them. The reduction in human labour and resources is achieved by the use of low cost components and very limited modification of the surroundings and the existing vehicles themselves. The operation of the cargo transit system has been verified and the relevant results are presented. An economical and robust cargo transit system is designed and implemented.**

## I. Introduction

When cargo is unloaded from ships, they will be placed on a transit vehicle which is automated. The cargo is then taken to the cargo storage location for unloading. Our system can automate the transfer of tons of cargo arriving at the ports every day. This would quicken the amount of cargo moving in and out of our country, thereby improving our economy. Time is money – this not a new insight. High-tech shipment sites such as CTA in the Port of Hamburg, Germany or the ECT in Rotterdam, Netherlands, [1] are impressive on account of their performance and make clear how efficient terminal design can provide new dimensions in cargo handling. Here automation is the only cost-effective solution. But in developing countries like India, this is not always the case. In numerous instances, automation which was successful in the developed countries could not be implemented in India because manual labour was much cheaper here and therefore autonomous solutions could not penetrate the Indian market. So, to implement an innovation here, the approach to our problem has to be cost centric as well appealing.

The market potential of an autonomous cargo transit system is huge considering the amount of cargo shipped is nearly 66 million tonnes every year. Most of the ports in our country have exceeded the capacity utilization rate, still the cargo moving in and out of India is ever increasing. Faster movement of cargo is needed and automation is an excellent way to achieve quick and efficient cargo transport. Most of the companies which provide autonomous solutions and services for cargo movement have their markets in the developed countries such as Germany.

Ports such as Hamburg in Germany where they have an automated material handling system is in place have cargo movement of 132.2 million tons every year. 28.642 million tons of cargo is shipped in and out of Tuticorin port every year and this number is ever increasing [1]. So, an autonomous solution has to be to be cost effective as well as being able to handle the massive amounts of cargo moving in and out every year.

## II. TECHNICAL BACKGROUND

The system of autonomous cargo transit in a port has been implemented in selected port terminals around the world. They specify that the surface of the port terminals be fitted with sensors, so the vehicles are able to navigate effectively. But, in developing countries such as India, there is minimal shortage of manual labour but an acute shortage of resources is present to implement such a complex system. Therefore, modifying the existing system to include sensors on the port would be a tedious and cost draining issue. The solution proposed would not need any changes to the existing system but only changes to the vehicles would be done which would fit into the budget constraints of developing countries.

The mapping of the terrain would be done with the help of an ultrasonic sensor attached to a servo motor which gives a 360 degree view of the entire area and monitor the position of each and every vehicle in the given area. The ultrasonic sensor is placed at a central location so that the entire range can be covered. Ultrasonic sensors work on a principle similar to radar, which evaluate attributes of a target by interpreting the echoes from radio or sound waves respectively. Active ultrasonic sensors generate high frequency sound waves and evaluate the echo which is received back by the sensor, measuring the time interval between sending the signal and receiving the echo to determine the distance to an object.

RF modules are used to form a mesh network which can communicate with the central hub of the port and transmit and receive data about the locations of the vehicles in the area. The algorithm is designed in such a way that the vehicles do not move out of the pre-specified area and remain in range of the ultrasonic as well as the RF modules. RF communication works by creating electromagnetic waves at a source and being able to pick up those electromagnetic waves at a particular destination. These electromagnetic waves travel through the air at near the speed of light. The wavelength of an electromagnetic signal is inversely proportional to the frequency [2].

Virtual nodes are made within the area which is in form of a square or a rectangle. The virtual nodes are created by subdividing the area into smaller fragments and each node is provided within a distinct address with the respect to the distance of the node from the origin. Each node is given an x and y- coordinate which is used by the vehicles to identify each node. The virtual node map is stored on each of the vehicles. This allows the vehicles to analyze where they are present in the area and quickly recalculate distances to their destinations rather than spending time on communication with the central hub

## III. PROBLEM STATEMENT

Cargo is unloaded from ships with the help of harbour cranes and is mounted on a transit vehicle in the material handling terminal of the port. The cargo has to be transported to the storage location where it can be dismantled or dispatched. The process of transporting the cargo is a highly manual and laborious task. It also requires large amount of planning and there is always a risk of befuddlement of cargo transportation, whereas automation is more effective and efficient than humans (e.g. manufacturing industry has become a boon with the advent of automation). Can automation be implemented in the container transport industries as well?

## IV. PROTOTYPE SPECIFICATIONS

For the development of the system's prototype, mapping a 2x2 meter squared terrain using an ultrasonic module (HC-SR 04) is done. The terrain is sub divided into various nodes and the node information is stored on each and every vehicle. The virtual nodes provide the waypoints for the vehicle when traversing through the area. Since practical implementation of the system in ports cannot be achieved at a prototype stage, the working model of the system is developed in a miniaturized form. The prototype will include two vehicles carrying loads and a central hub which will be the Intel Galileo.

The Intel Galileo would act as the central hub (control unit) for the entire system. It manages and controls the motion of all the vehicles with the help of RF communication and Ultrasonic mapping. The peripherals of the Intel Galileo used in the system will be: SPI, I2C, RS-232 serial port, Micro-SD, I/O ports for integrating with communication module.

## V. IMPLEMENTATION

### A. Hardware Implementation:

The Processor (Intel Galileo board) will be central hub of the entire system. It manages various tasks including RF communication, ultrasonic ranging and controlling the swarm vehicles. A full-sized mini-PCI Express slot, 100 Mb Ethernet port, Micro-SD slot, RS-232 serial port, USB host port, USB client port, and 8 Mbyte NOR Flash come standard on the board. In this system, the features which will be used and implemented are SPI, I2C, Micro-SD slot, RS-232 serial port and the NOR Flash.

The ultrasonic module [3] will map the 2x2 terrain which will be similar to a port terminal. The range of the ultrasonic module is 4m, so mapping a 2x2 terrain should be significantly accurate. The 360 degrees map of the terrain is achieved by attaching a servo motor onto the ultrasonic module, which constantly rotates back and forth at constant speed. The ultrasonic module retrieves vital information about the positions from terminal vehicles or obstacles inside the terrain. These data is sent to the central unit from there to a remote PC through RS-232 serial port communication. It is also stored on a Micro-SD for further analysis. The data obtained on the PC will be displayed as Graphical User Interface (GUI) on RADAR using "Processing" software. The central unit also sends the mapped data through RF communication [2] to the individual terminal vehicles.

The RF module used here is nRF24l01+. The nRF24L01 is a highly integrated, ultra low power (ULP) 2Mbps RF transceiver IC for the 2.4GHz ISM band. It has a 1.9 to 3.6V power supply range, 1Mbps and 2Mbps on-air data rate. It can be tuned to 128 different frequencies which allows for the control of the same number of terminal vehicles. Therefore, it is able to control as much as 128 vehicles within the specified area and the high data rate allows for quick and efficient communication. The coordinates of the destination and virtual nodes of the vehicles are sent via the RF module attached to each vehicle and the control unit. Each vehicle [4] is tuned to different frequencies within the 2.4GHz band so that it receives the correct destination. The frequency tuning of the RF module is done by the control unit so that locations can be transmitted to each vehicle. A microcontroller (ATMEGA 8) is placed on each vehicle to decode the RF data received [2].

The ATMEGA 8 placed on each vehicle controls the RF module [2], motor drivers and motor control/movement. Whenever a load or container has been placed on the vehicle, a switch is pressed which activates the vehicle. As soon as the vehicle has been activated, it receives the coordinates of the destination and the motor controls are activated. Once the container has been unloaded once it has reached the destination, the vehicle retraces its path back to its terminal port for its next cycle of load. The path followed by the vehicle to the destination in the form of virtual nodes is stored in a hash table so that the vehicle can retrace its path back to the terminal.

**B. Software Implementation:**

The IDE used for software development is Arduino 1.5.3, which has support for Intel Galileo since it uses an x86 processor. The Arduino development environment contains a text editor for writing code, a message area, a text console, a toolbar with buttons for common functions, and a series of menus. It connects to the Arduino hardware to upload programs and communicate with them. The raw data from the ultrasonic module is sent via serial port to software called "Processing".

It serves as a software sketchbook to create computer programs within a visual context. The data received via serial port is made into useful information and its object variable (vehicles position) is processed by various functions written in java programming language to create a standalone application. The standalone application consists of a RADAR structured 2D-canvas drawn using various written functions which provide visual display of entire terrain. The distance and the angle of each vehicle [4] with respect to the origin is also displayed in the canvas. (Figure 1)

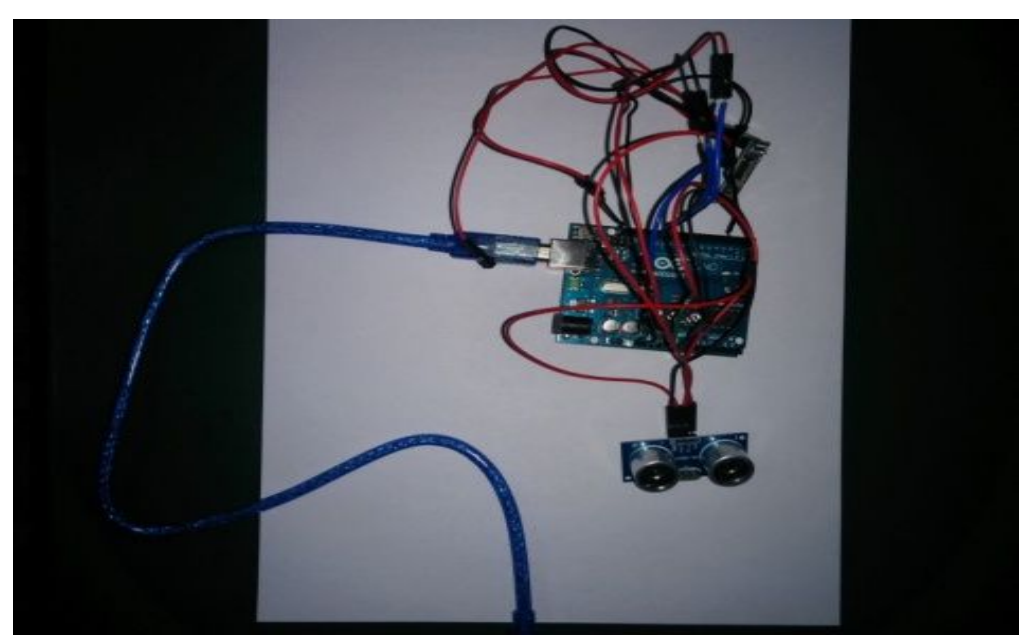

Fig. 1- Arduino connections with Ultrasonic and RF modules

The vehicles have to avoid colliding with each other [5] and also going out of RADAR. For achieving this, the best out of the available shortest path algorithm was implemented. Some of the shortest path algorithms are Bellman-ford algorithm, Dijkstra's algorithm, A* algorithm, Floyd Warshall algorithm [6]. The general working of this algorithm depends on the various nodes between the source and the destination. These algorithms are derived from various algorithm design procedure such as Dynamic programming, Amortized analysis of various data structures and Graph theory. Graph theory is implemented to determine various sets of vertices and edges onto a binary tree which has a weighted variable to determine the shortest distance path [6]. Virtual nodes are pre programmed onto the microcontroller of the vehicle. Virtual nodes have to be assigned to each vehicle so that collision avoidance is possible [5].

The velocity of each vehicle [4] will be the same and fixed and also the distance between two nodes is known from which the time of rotation of motor can be specified. The velocity of the vehicle can be maintained constant by the use of PID controller and the orientation of the vehicle will be pre programmed since the specification of the nodes will be known.

The pins of the Ultrasonic module (Figure 2) Echo, Trigger are connected to the I/O Arduino Uno (Figure 3), Vcc to the 5V pin on Arduino and the Gnd to the Gnd of the Arduino. Arduino IDE 1.0.5 was used to compile and run code on the microcontroller in the Arduino. The servo motor (Figure 4) is attached to the Arduino whose speed of rotation is in sync with the RADAR motion obtained using "Processing".

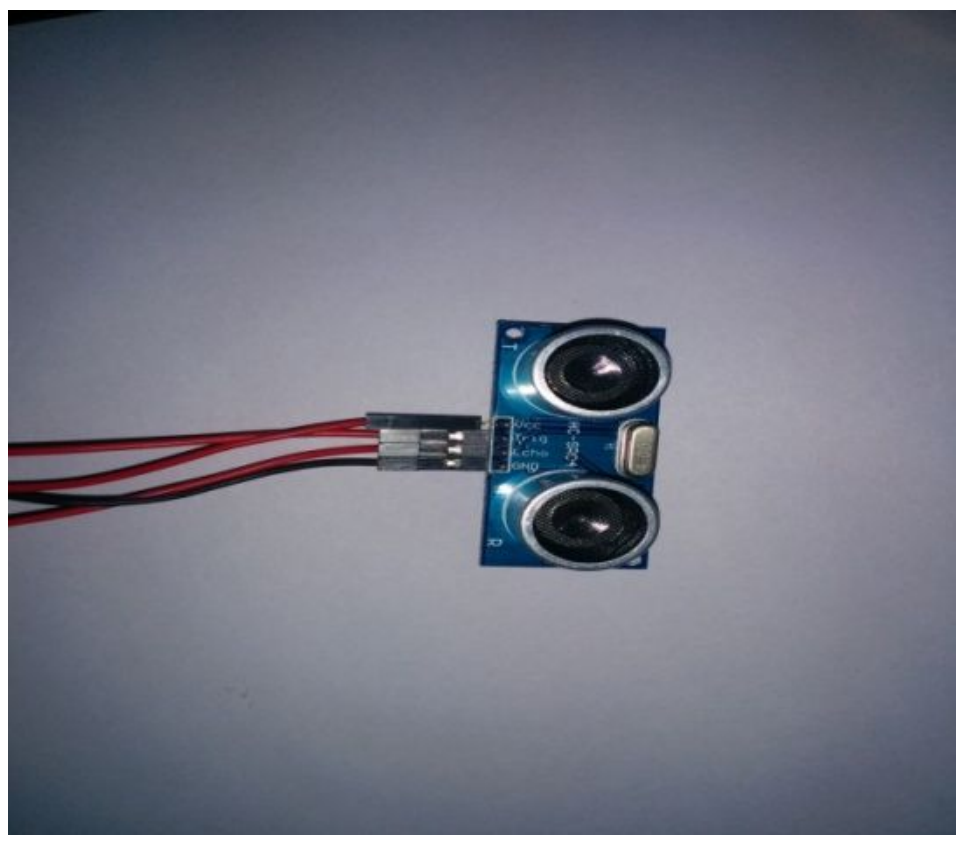

Fig 2 - Ultrasonic Module (HC SR-04)

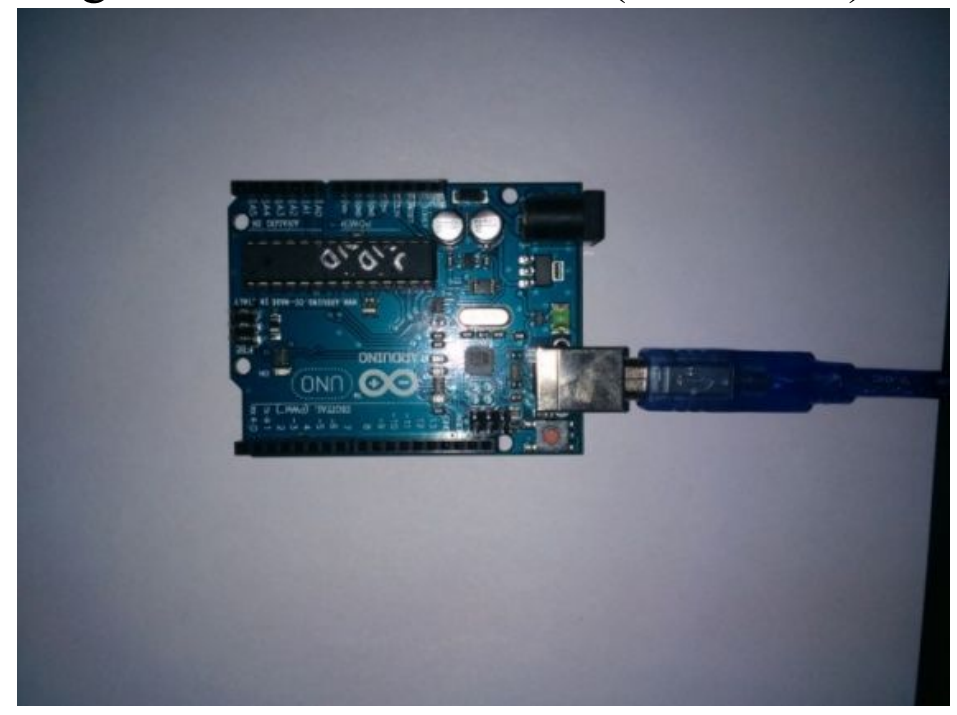

Fig 3 - Arduino Uno Board

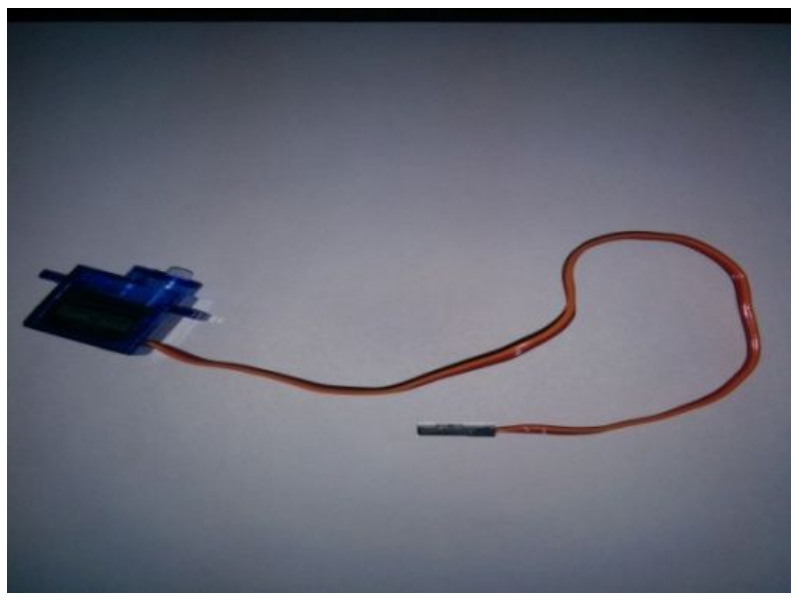

Fig 4 - Servo Motor

The ultrasonic data is sent via the serial port to the PC which is loaded with the software "Processing". Processing gives an output GUI in the form of a 2D RADAR screen (Figure 5). RF (Figure 6) communication between two Arduino's was worked out first (Figure 7). Then, the location of the vehicle from the Ultrasonic map from one Arduino was sent to the other Arduino via RF communication (Figure 3). It was also verified using the serial monitor of the Arduino.

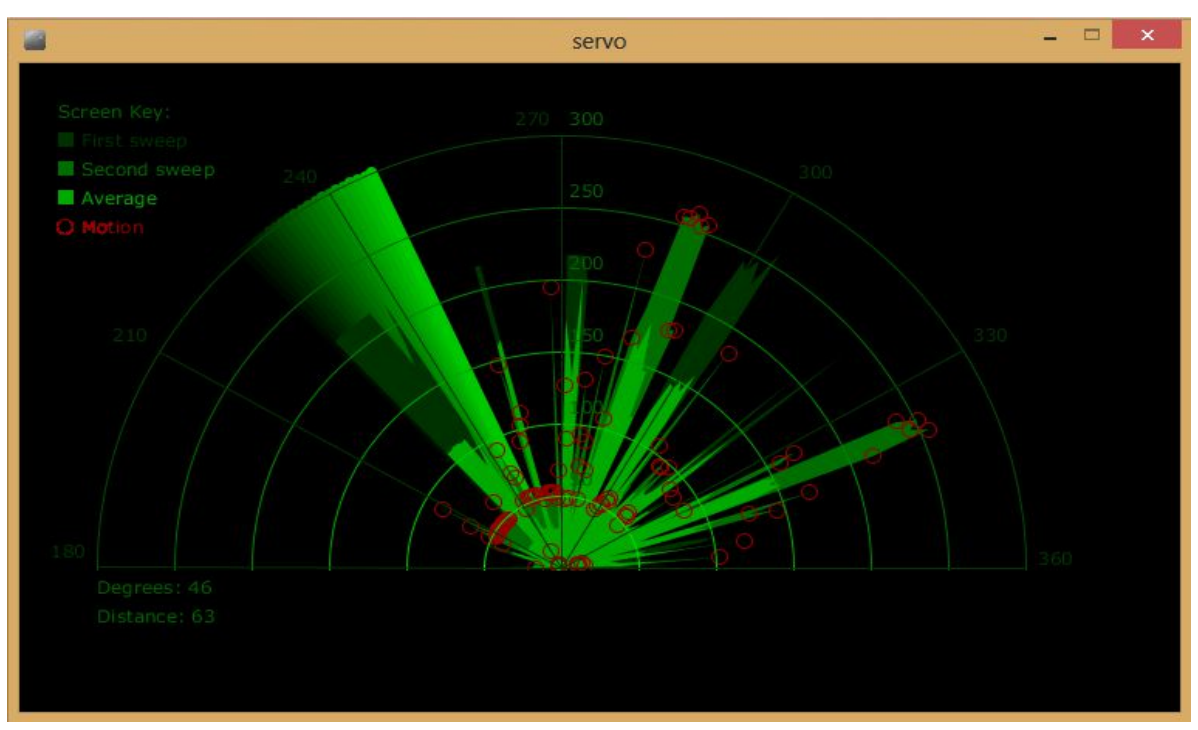

Fig 5 - RADAR using Processing

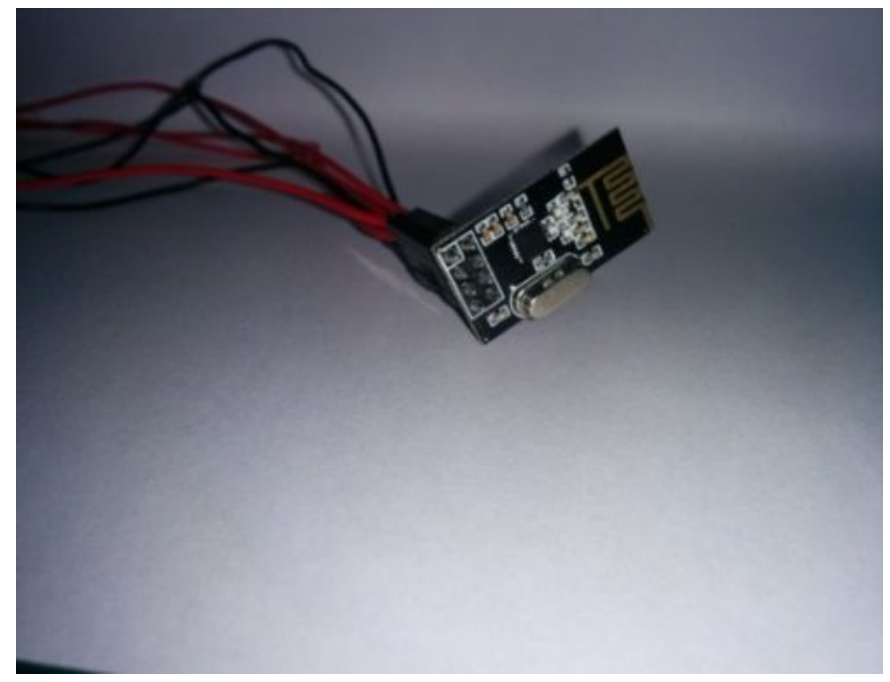

Fig 6 - RF Module (nRF24l01+)

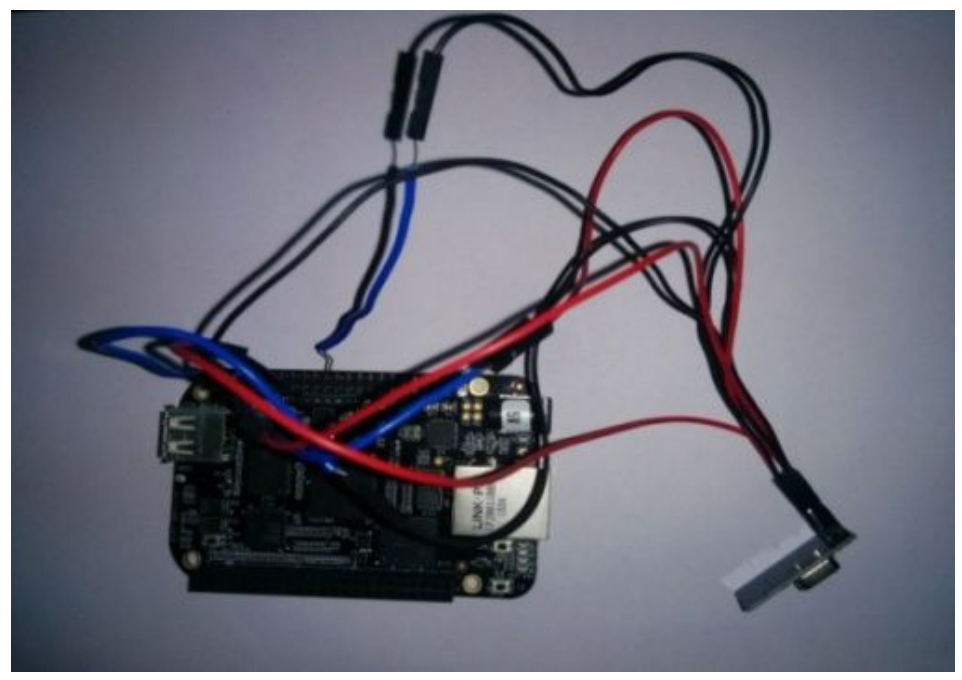

Fig 7-BeagleBone using Arduino Libraries

**C. Mechanical Implementation:**

The dimensions of the vehicle

1. Wheel base -360 mm
2. Track width -350 mm
3. Load placement area - 410 * 250 $mm^2$
4. Diameter of the wheel -100 mm

The vehicle [4] should be able to withstand the total weight of the cargo. So the material must have high yield strength to avoid fracture. Since the vehicles used being scaled down versions of the real time vehicles, a relatively softer material will be used. Therefore the chassis of the vehicles were made of mild steel, but when it comes to real time applications, stronger materials have to be considered.

The prototype has three wheels, two driven in the rear and one wheel in the front. The two wheels in the rear are driven by the motors and the same controls the direction as well. This vehicle was designed with the idea of weight reduction in mind. This was the reason the concept of one wheel in the front was formulated and subsequent simulations proved it effective. But when comes to real time application in the port or any real life scenario, this three wheel design will not be able to withstand the load. Therefore, the four wheels design would be suitable in those cases. A small compartment is attached below the platform to fit the electronic components such as micro controller and the drivers and will also be unaffected by load being place on top of the vehicle.

The mechanical design of the vehicle has also been determined which has been described in the mechanical implementation part and the 3D design of the vehicle in CREO has also been included in the appendix (Figure 8,9).

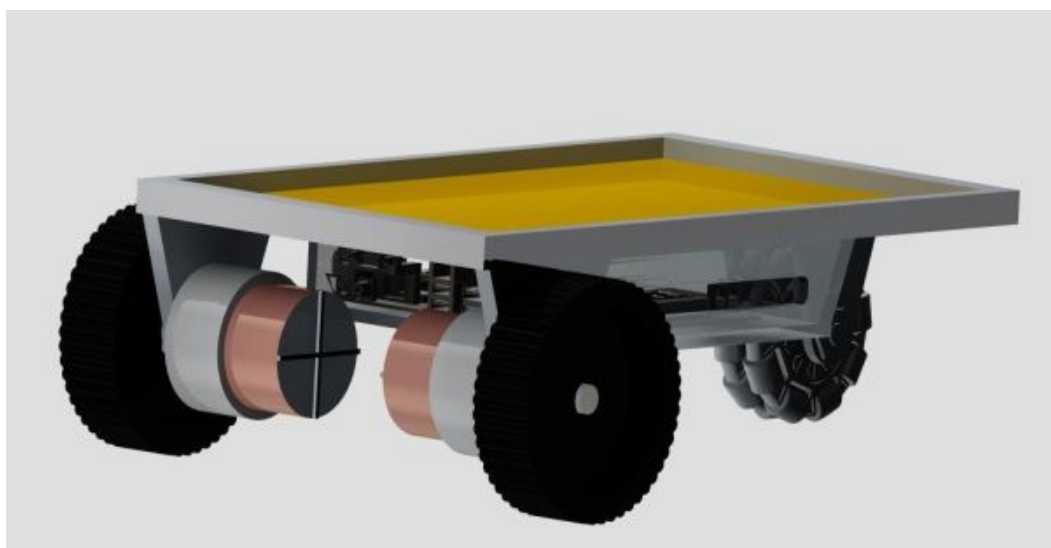

Fig 8 - 3D Model of a swarm transit vehicle designed in Creo Parametric 1.0

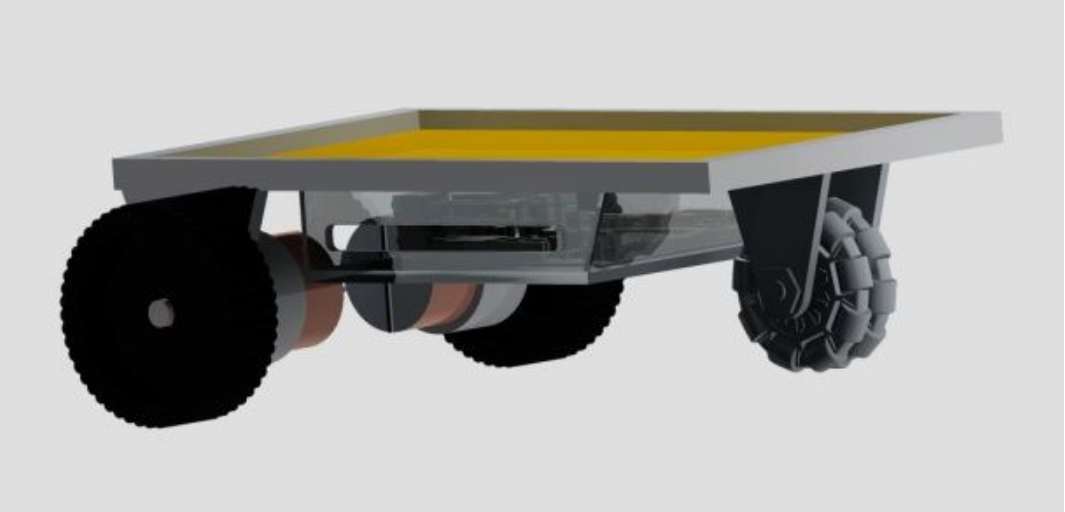

Fig 9 - 3D Model of a swarm transit vehicle designed in Creo Parametric 1.0

## V. RESULTS

Real time data is obtained from the vehicles regarding the speed, distance moved and angle with respect to both the x and y axis. The data is then sent to the central hub and the results are plotted and can be used for monitoring and maintenance purposes. Several trials were carried out with the vehicles and real time data has been collected. The data is then plotted against each and every parameter to analyze the parameters of each vehicle during transit. The distance moved by each vehicle along both the x-axis and y-axis with respect to time is plotted which gives us the resultant velocity of each vehicle in real time. The velocity data collected is monitored by people for any performance issues or any defects. The distance travelled from the origin is plotted with the respect to the angle of each vehicle at present time to indicate and identify the error factor and correct it in the next iteration. The logging data is stored with the Intel IoT analytics. The data obtained is also accessible via the Intel IoT Analytics dashboard. This is done by logging all the data obtained from the vehicles onto the Intel Galileo which is added to the Intel IoT using the internet

**Table 1. Components Needed/Used for the Prototype:**

| Item | Number | Cost |
|---|---|---|
| **Intel Galileo** | 1 | 6000 |
| **HC-SR04** | 2 | 200 |
| **Servo Motor** | 1 | 200 |
| **RF Module (rf24l01+)** | 5 | 1000 |
| **Intel 8051** | 4 | 400 |
| **Wheels** | 4 | 100 |
| **Motor Driver IC** | 4 | 400 |
| **Motors** | 8 | 1600 |
| **Omni Wheels** | 3 | 2100 |
| **Chassis** | 4 | 200 |
| **Miscellaneous** | 1 | 500 |

The total cost to make the prototype was 12,700 Indian Rupees. The automation of the terminal ports will greatly increase the amount of cargo entering and leaving the harbour. There will be increased amount of export of goods from our country to the world, (e.g. The Port of Hamburg, Germany increased in exports and imports after autonomous system was in place) thereby improving our economy and the trade industry.

## VI. CONCLUSION

The miniaturized version of the port automation can be implemented in a port with the help of port authorities. This would require larger vehicles which would have to be manufactured and would require four wheel drive instead of a three wheel system. The algorithms needed to develop such a system would have modified according to the system and the port. The sensors would have to be altered if the system is to be implemented in a real port. Mapping systems with a range of 1-5 km will have to be designed which will need to have a high accuracy. The harbor cranes have to be recalibrated to fit the specifications of the automated port solutions.

## VIII. APPENDIX

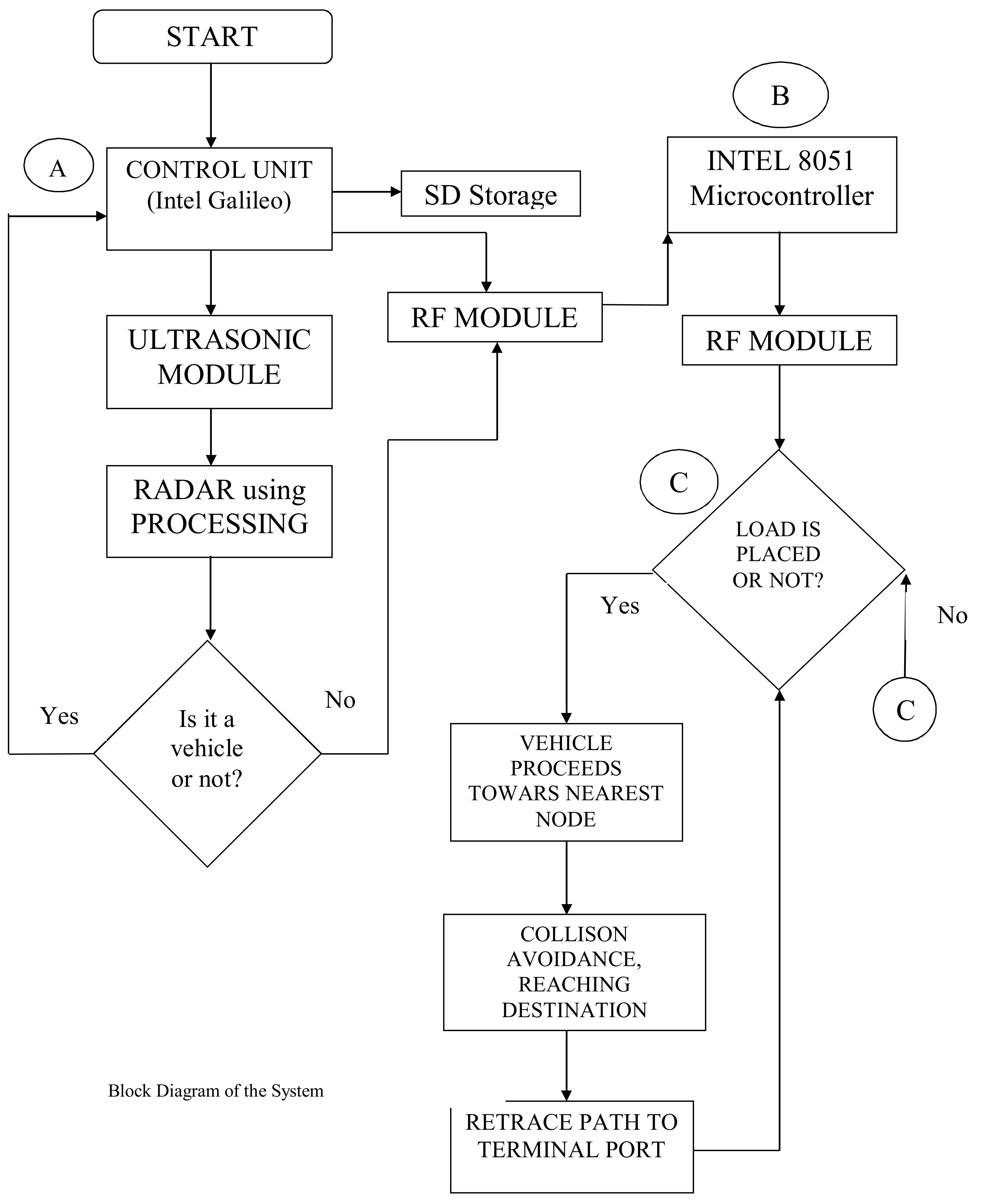

Block Diagram of the System